\documentclass[lettersize,journal]{IEEEtran}
\usepackage{amsmath,amsfonts}
\usepackage{algorithmic}
\usepackage{algorithm}
\usepackage{array}
\usepackage[caption=false,font=normalsize,labelfont=sf,textfont=sf]{subfig}
\usepackage{textcomp}
\usepackage{stfloats}
\usepackage{url}
\usepackage{verbatim}
\usepackage{graphicx}
\usepackage{multirow}
\usepackage{cite}
\usepackage{pifont}
\usepackage{nccmath}
\usepackage[table]{xcolor}
\newcommand{\cmark}{\ding{51}}%
\newcommand{\xmark}{\ding{55}}%
\usepackage{amsthm}
\usepackage[shortlabels]{enumitem}
\usepackage{listings}
\usepackage{framed}
\newtheorem{definition}{Definition}

\begin{document}
\lstset{language=Python}

\title{Signature and Log-signature for the Study of Empirical Distributions Generated with GANs}

\author{Joaquim {de Curt\`o}\textsuperscript{\textsection},~\IEEEmembership{Member,~IEEE,} \and Irene {de Zarz\`a}\textsuperscript{\textsection},~\IEEEmembership{Member,~IEEE,}\\ \and Hong Yan,~\IEEEmembership{Fellow,~IEEE,} \and Carlos T. Calafate,~\IEEEmembership{Member,~IEEE}.
        
\thanks{J. de Curt\`o and I. de Zarz\`a are with Centre for Intelligent Multidimensional Data Analysis, HK Science Park, Hong Kong, with Universitat Polit\`ecnica de Val\`encia, Val\`encia, and also with Universitat Oberta de Catalunya, Barcelona (e-mail: \{decurto,dezarza\}@uoc.edu).}

\thanks{H. Yan is with Centre for Intelligent Multidimensional Data Analysis, HK Science Park, Hong Kong, and also with the Department of Electrical Engineering, City University of Hong Kong, Kowloon, Hong Kong (e-mail: h.yan@cityu.edu.hk)}

\thanks{C. T. Calafate is with Universitat Polit\`ecnica de Val\`encia, Val\`encia (e-mail: calafate@disca.upv.es).}

\thanks{This work is supported by HK Innovation and Technology Commission (InnoHK Project CIMDA) and HK Research Grants Council (Project CityU 11204821). Thank you to Terry Lyons for useful discussions about the Signature Transform and the CIMDA-OXFORD group for great insights on the topic.}

}

\markboth{De Curt\`o et al.\ -- Signature and log-signature for the Study of Empirical Distributions Generated with GANs}{De Curt\`o et al.\ -- Signature and log-signature for the Study of Empirical Distributions Generated with GANs}

\maketitle
\begingroup\renewcommand\thefootnote{\textsection}
\footnotetext{Equal contribution.}
\endgroup

\begin{abstract}
In this paper, we bring forward the use of the recently developed Signature Transform as a way to measure the similarity between image distributions and provide detailed acquaintance and extensive evaluations. We are the first to pioneer RMSE and MAE Signature, along with log-signature as an alternative to measure GAN convergence, a problem that has been extensively studied. We are also forerunners to introduce analytical measures based on statistics to study the goodness of fit of the GAN sample distribution that are both efficient and effective. Current GAN measures involve lots of computation normally done at the GPU and are very time consuming. In contrast, we diminish the computation time to the order of seconds and computation is done at the CPU achieving the same level of goodness. Lastly, a PCA adaptive t-SNE approach, which is novel in this context, is also proposed for data visualization.
\end{abstract}


\section{Introduction}
\IEEEPARstart{C}{omputing} in a reasonable amount of time a metric to accurately assess the quality of synthetic samples produced by Generative Adversarial Networks (GAN) has been at the core of research in Computer Vision since its first appearance in \cite{Goodfellow14}. Yet being FID \cite{Heusel17} the most common measure, it relies on an inception module that requires an intensive GPU usage and takes a lot of time to be calculated for only one epoch of the generated data.
\\

In this work we propose a two-fold approach. On the one hand, we present a score function on top of the Signature Transform \cite{Lyons2014} to assess image quality in an unprecedented way; reliable, fast and easy to compute epoch by epoch. On the other hand, we use techniques from statistics to study the goodness of fit of the generated distribution, given a standardized pipeline for the interpretation of results of the converged sample distribution. Statistical techniques are one-liners that can be computed on-the-fly with no computation overhead whatsoever.
\\

As a case study where lots of data are available, huge funding opportunities are arising, and synthetic image generation could be of use, we will deviate first our course of explanation and focus on data extracted from planetary missions, both from mobile platforms such as a rover or a base station, as well as Unmanned Aerial Vehicles traversing the surface of other planets. In the most generic example of vision, we will first collect data, visualize it through k-means clustering, together with a new variant of t-SNE (t-distributed Stochastic Neighbor Embedding), and we then proceed with synthetic image generation and the extraction of meaningful information. A pipeline to study the generated data using statistical analysis is presented, as well as, a primer on the Signature Transform that serves as an introduction prior to the presentation of the new measures.
\\

Space Exploration has become ubiquitous, from NASA missions to other planets, to the privatization of the space sector led by companies such as SpaceX, Blue Origin or Virgin Galactic. The need to study and analyze the amount of data that will be generated in such missions will become essential to improve the Return on Investment. Moreover, the current trend of sending small unmanned autonomous vehicles (nanorovers) to the Moon (e.g. Astrobotic), and in the future also to Mars and other planets, will make research in this setting available to pursue at universities worldwide. In this context, our aim is to provide a first approach to deal with the data collected by NASA Perseverance, and to extract purposeful information. We first visualize the data, for instance by the use of k-means clustering and t-SNE, and then extract a subset of the samples useful for the application under consideration. In our case we focus on synthetic image generation, and we then finally go through thorough testing, analysis and review of the algorithms used and the results obtained.
\\

The manuscript is organized as follows. Section \ref{sn:overview} gives an introductory overview to the manuscript and a motivating usage case. The process of data collection and visualization is then addressed in Sections \ref{sn:collection} and \ref{sn:tsne}, respectively. Section \ref{sn:syntheticsamples} explains thoroughly the process of synthetic data generation. A pipeline to study the synthetic distributions by the use of statistical techniques is provided in Section \ref{sn:statistical_analysis}. The metrics based on the Signature Transform are developed in Section \ref{sn:signature}, along with detailed experimentation and plots of the spectrum. Visual exploration of the data using PCA Adaptive t-SNE is given in Section \ref{sn:exploration}. Finally conclusions are yielded in Section \ref{sn:conclusions}.

\section{Overview}
\label{sn:overview}

With the advent of Deep Learning, applications that rely on huge amounts of data have emerged to be game changers in a wide range of topics and across the disciplines \cite{Girshick14,He15,Girshick15,Long15,Redmon16,Gatys16,Gatys16_2,He17,Vaswani2017,Wang18_2,Karras18,DeZarza22}. The dramatic improvement in accuracy and speed has paved the way for the first time to use automated learning techniques in scenarios where reliability is key, such as safety-critical systems and self-driving cars \cite{Chen17,Chen17_3,Yang18,Parmar18,Brock19,Mildenhall2020,Park20}. The use of these techniques in space exploration is starting to take place; as conditions in other planets are adverse for humans, robots are in charge of teleoperated missions on their own prior to human settlement. Autonomy will be crucial for the possibility of humanity to establish elsewhere.
\\

In this work we initiate a journey through several techniques that will give us the ability to understand the data, analyse it and even generate new synthetic samples, sometimes indistinguishable at the human eye, by learning the original distribution by the use of examples, and then sampling from it. We will also extract semantic information from the collected images as a way to show that we can successfully apply the same techniques we use on Earth on other planets for systems that learn by themselves. These algorithms are the seed for more complex enterprises under the condition of scarce data. After all, our ability to compile large amounts of information from other planets is bounded, such as SLAM and VIO.
\\

The field of synthetic image generation has seen rapid progress. The necessity to generate synthetic imagery given some training data in many applications (simulated environments, additional training data, style-transfer) has seen great research efforts to establish a stable, principled way to accomplish the task. Generative Adversarial Networks (GANs) \cite{Goodfellow14,Odena17,Antoniou18,Salimans16,Mescheder17,Mescheder18,Jolicoeur2019,Karras2019,Karras2021} and VAE (Variational AutoEncoders) \cite{Kingma14} offer stable training mechanisms to achieve convergence. Yet, more progress is needed as the capacity of the network is mainly bounded by the GPU memory and training resources available \cite{Chen17_3,Dosovitskiy16,Zhao17,Karras18,Wei18,Brock19}. Results often suffer from mode collapse and gradient explosion, and its good performance to accomplish complex tasks such as generating additional multi-view frames is yet to be proven. 
\\

The work presented in \cite{Song2019} introduced a new type of generative model based on annealed Langevin \cite{Roberts1996,Welling2011}. The method is further developed in \cite{Song2020}, where they show competitive image generation. Diffusion Probabilistic Models \cite{Ho2020} achieved state-of-the-art results on CIFAR10 building on the same principles derived from diffusion-based methods \cite{Goyal2017}. However, Score-based Generative Models \cite{Jolicoeur2021} suffer mainly from the same drawbacks of GANs and their real-time implementation is not viable due to the sampling step where the dimension of the output must be the same as the dimension of the input. That is, they are hugely dependent on GPU memory resources and necessitate high computing time.
\\

Supplemental recent approaches \cite{Zhao2021} are based on the attention mechanism \cite{Vaswani2017} building mainly on Vision Transformers \cite{Dosovitskiy2020}. Other techniques like NeRF \cite{Mildenhall2020} could be essential to add structure to the learning paradigm. In the context of space exploration, these techniques could have a huge impact, as the number of data is by definition scarce; although diminishing, due to our limitation to collect high-resolution data in other planets. Bandwidth and latency of operating a commercial robot at thousands of km from Earth severely limit the real-time data acquisition and capacity of reaction of human-controlled rovers. Indeed, being able to synthesize new useful data from few observations could enable fast and improved SLAM and VIO, as well as enhance autonomous capabilities of the extraterrestrial robot counterparts. 

\section{Data Collection}
\label{sn:collection}

The first stage of our study consists on data collection. We provide scripts to massively scrape data from NASA Website\footnote{https://mars.nasa.gov/} 
using the tools Selenium\footnote{https://www.selenium.dev/} 
and Beautiful Soup\footnote{https://www.crummy.com/software/BeautifulSoup/}. These tools allow the user to setup an agent to automatically navigate using the browser a given webpage (static or dynamic) with the purpose of data collection. For instance, in Listing \ref{vm1} we show a pseudo-code example of an agent automatically collecting data from a website: loads the browser, navigates to a site, scrolls the window to have the target header in view and then extracts the source for later processing.\\

\begin{lstlisting}[label=vm1,caption=Example of an agent automatically collecting data from a website,float,frame=tb]
# Start session in browser.
driver = webdriver.Browser()
driver.implicitly_wait(30)
driver.maximize_window()

# Navigate to given site.
driver.get(site[0])

# Slightly scroll the window 
# to have header in view.
driver.execute_script("window.scrollTo(0,400)") 

# Extract source.
page_source = driver.page_source
soup = BeautifulSoup(page_source,'lxml')
\end{lstlisting}

\section{Visualization}
\label{sn:tsne}
\begin{figure*}[ht]
\centering
\includegraphics[scale=0.37]{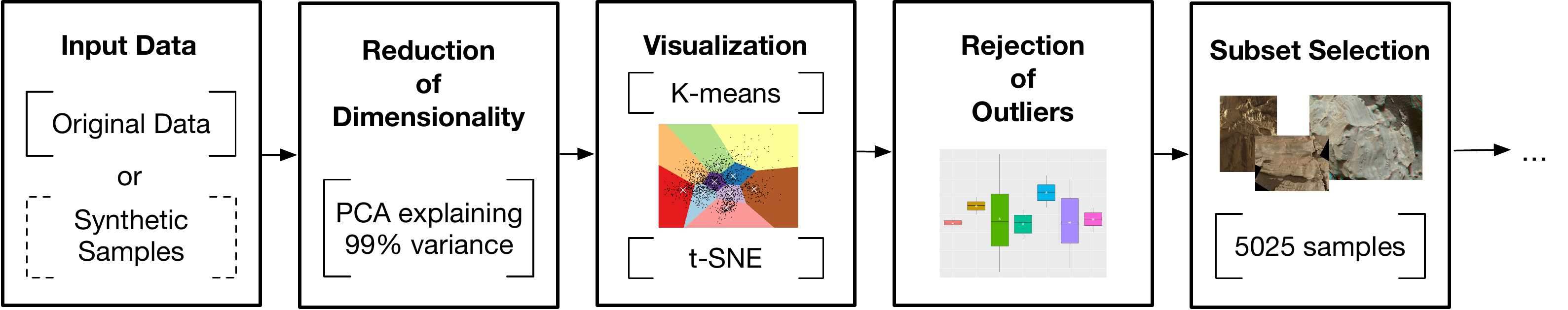}
\caption{Visualization pipeline under study.}
\label{fge:09_decurto_and_dezarza}
\end{figure*}

With the aim of using a subset of the collected data for synthetic image generation, we visualize the data to look for a significant portion of the samples to use, and identify outliers and points outside the distribution. Also, it is important to consider that images are taken using contrasting configurations of cameras; we choose a subset that describes well the terrain of the surface of Mars. The techniques under study are k-means clustering with a prior projection on the 2D plane by the use of PCA($2$) and t-SNE, with a prior reduction of dimensionality using PCA which adjusts the number of Principal Components to explain for 99\% of variance, as shown in Figure \ref{fge:09_decurto_and_dezarza}.

\subsection{K-means Clustering}
We project the data into a 2D plane by the use of PCA, and then apply k-means clustering, where $k$ is the number of cameras used on each mission. Due to the high dimensionality of the problem, the need to use a more sophisticated technique arises. The resultant projection does not show well defined clusters, and not all outliers are correctly rejected. K-means clustering is a lineal technique and a lot of information is lost on the 2D plane projection. So, we then propose to use t-SNE, which is based on a probabilistic measure, to accomplish the task. We introduce an adaptive t-SNE technique with a varying degree of explainability of variance given by PCA.

\subsection{t-SNE}
\begin{figure}[ht]
\centering
\subfloat[]{\includegraphics[scale=0.13]{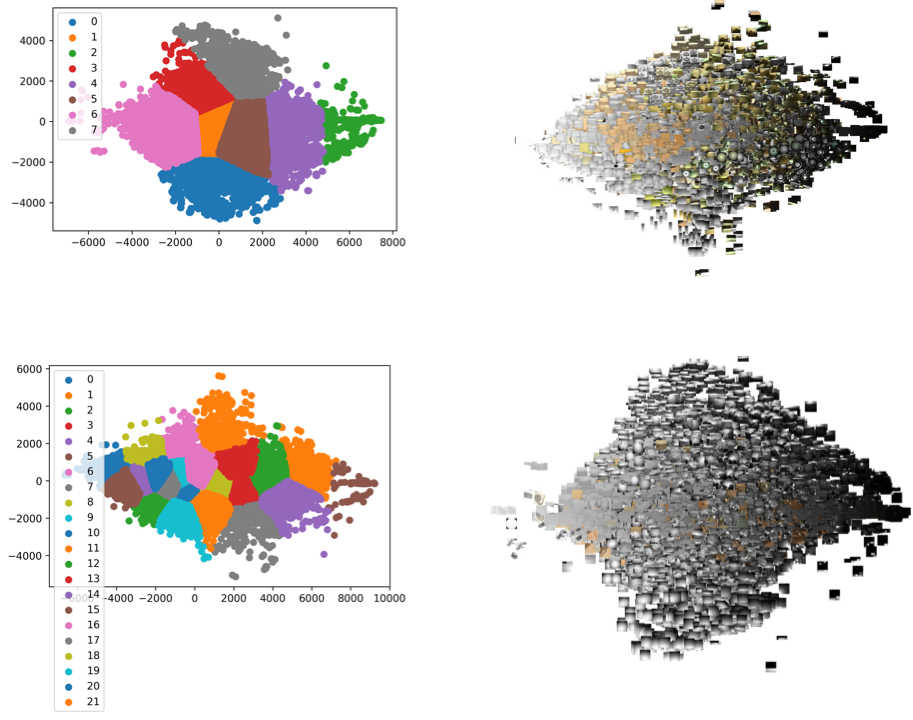}}
\subfloat[]{\includegraphics[scale=0.11]{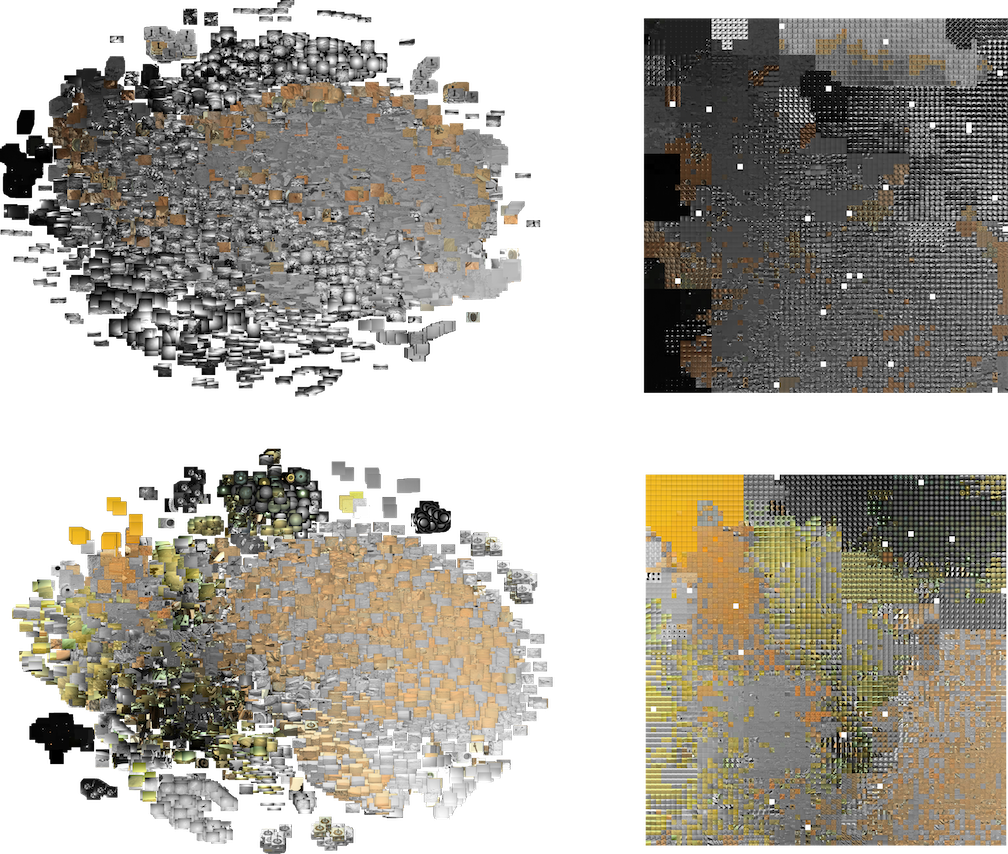}}
\caption{K-means Clustering (left) and t-SNE (right) on images from NASA Curiosity (upper figures) and Perseverance (lower figures).}
\label{fge:01_decurto_and_dezarza}
\end{figure}

t-SNE (t-Distributed Stochastic Neighbor Embedding) was proposed in \cite{vandermaaten08a}, and it introduces a probabilistic technique to visualize and understand high-dimensional data such as images. We perform a prior dimensionality reduction by the use of PCA, in which we choose adaptively the number of Principal Components that explains the 99\% of the variance, as we can see in Figure \ref{fge:01_decurto_and_dezarza}; we name it PCA adaptive t-SNE. Having this adaptive behavior is central to the task at hand to extend the validity of the technique to a varying number of data points, camera settings, and changing environments.

This technique allows us to select a subset of the original data taken by MastCam-Z, which is useful for synthetic terrain generation and instance segmentation. It helps identify out-of-distribution points, and clusters together all images pertaining to cameras pointing to martian terrain. The subset of data is released as a dataset 
on its own, both in PNG image format and in TFRecords for rapid importation and training.

\section{Synthetic Generation of Samples}
\label{sn:syntheticsamples}

We focus on the problem of generating synthetic images with a limited amount of data, being Stylegan2-ADA \cite{Karras2020} the baseline method of choice for our studies.

\subsection{Stylegan2-ADA}
\begin{figure}[ht]
\centering
\includegraphics[scale=0.24]{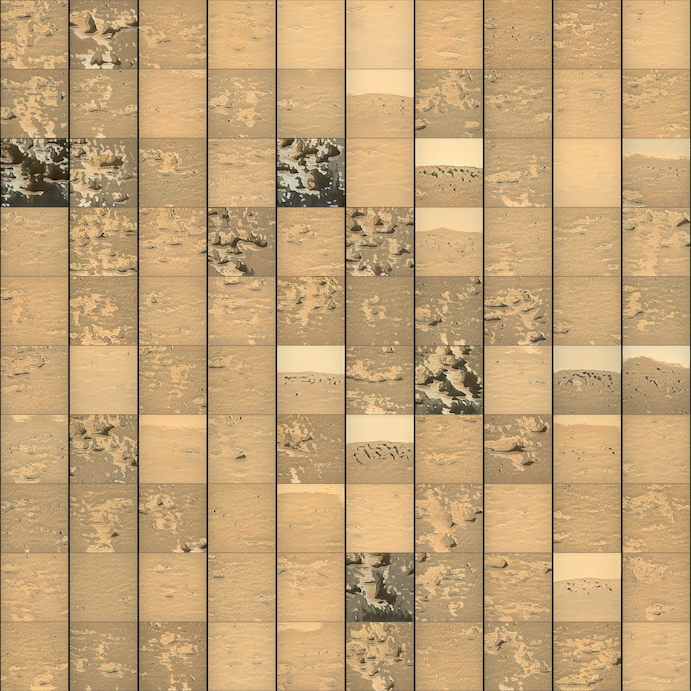}
\caption{Grid of $100$ samples generated by Stylegan2-ADA trained at size $256\times256$ using samples from NASA Perseverance.}
\label{fge:03_decurto_and_dezarza}
\end{figure}

GANs learn a probability distribution from samples by training concomitantly two networks, where the Generator (G) produces images that resemble the original training instances, while the Discriminator (D) determines their fidelity. The networks are trained until convergence in a zero-sum game fashion. In \cite{Karras2020} the authors go beyond the common GAN architecture by leveraging the concept of Stochastic Discriminator Augmentation, and proposing an Adaptive Discriminator Augmentation (ADA) that helps the network converge to the same accuracy levels as before, but with a few thousand samples; which is particularly suited for our application. We train the network with a subset of $5025$ samples from the NASA Perseverance mission using an NVIDIA-P100 on the cloud during 48h. The results obtained are consistent. We sample the trained model to generate $100$, $1000$ and $10000$ samples, as shown in Figure \ref{fge:03_decurto_and_dezarza}, and release the data publicly for testing purposes.

\subsection{Score-based Generative Modeling}
Based on annealed langevin \cite{Song2019,Song2020}, Score-based Generative Modeling introduces a new way to generate synthetic data as an alternative to adversarial learning in GANs.
\\

\begin{figure}[ht]
\centering
\subfloat[]{\includegraphics[scale=0.24]{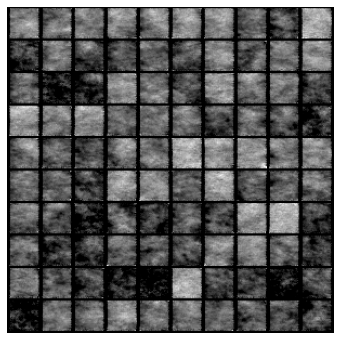}}
\subfloat[]{\includegraphics[scale=0.24]{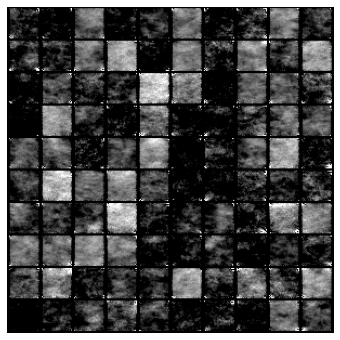}}
\subfloat[]{\includegraphics[scale=0.24]{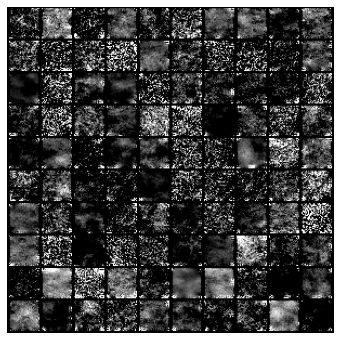}}
\caption{Grid of $100$ samples generated by PC Sampling (left), Euler-Maruyama (middle) and ODE Sampling (right). The models are trained at size $28\times28$ during $71$ epochs using samples from NASA Perseverance.}
\label{fge:17_decurto_and_dezarza}
\end{figure}

We conduct synthetic image generation using grayscale data from NASA Perseverance at image size $28\times28$, and sample the learned distribution using PC (Predict-Correct) Sampling, Euler-Maruyama, and ODE Sampling, see Figure \ref{fge:17_decurto_and_dezarza}.
\\

If we then compute the likelihood on the dataset by the use of the learned model and a batch size of $32$ we find an average number of bits per dimension of $6.18$; for comparison purposes, notice that the average number of bits per dimension on MNIST is $3.98$. This means the subset of samples from NASA Perseverance we have chosen is a good candidate to test learning models such as GANs and Score-based Generative Modeling because the learned latent spaces have to incorporate complex features such as description of the terrain, rocks and sky, while keeping the number of samples low. 

\section{Statistical Analysis of the Generated Distribution}
\label{sn:statistical_analysis}

\begin{figure*}[ht]
\centering
\includegraphics[scale=0.4]{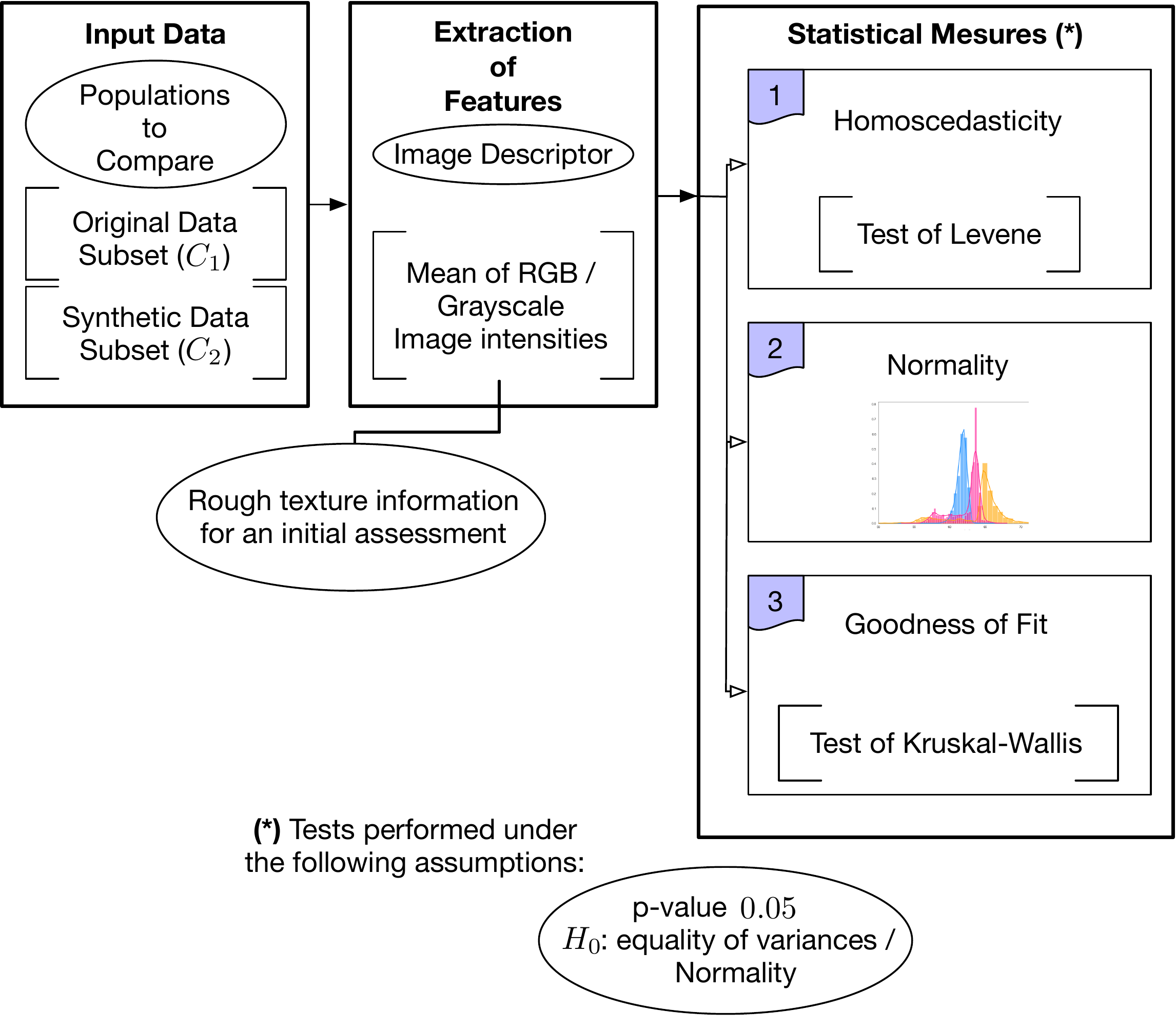}
\caption{Statistical measures to study goodness of fit to initially assess GAN convergence.}
\label{fge:14_decurto_and_dezarza}
\end{figure*}

Considering the original subset of data from NASA Perseverance, and the generated synthetic dataset of $1000$ images as two distributions we want to compare, we propose a preliminary statistical analysis by means of a non-parametric test: Kruskal-Wallis \cite{KruskalandWallis52}. In order to proceed with the evaluation, we compute the mean of the RGB image intensities or gray-scale as a proxy of image descriptor; although undecorated, mean intensity can provide rough texture information for an initial assessment. We first test the homoscedasticity, or equality of variances, by means of the test of Levene; and we also test for normality of the two distributions. Finally we compute goodness of fit by Kruskal-Wallis. In the case of homoscedasticity we reject the null hypothesis (significant $p$-value less than 0.05), and the same occurs for the test of normality of the original distribution (as expected as the original samples do not follow a distribution Gaussian). In the case of the synthetic samples, we accept the null hypothesis for the case of normality, and we can assure the distribution is normal, which makes sense as the GAN architecture initially models the samples as White Gaussian, and then modifies them step by step to fit the original distribution. Nevertheless, we cannot accept the null hypothesis for goodness of fit, which means that a more sophisticated way of measuring the sample quality in GAN has to be proposed, as has already been extensively seen in the current literature. For example, MS-SSIM \cite{Odena17} and FID \cite{Heusel17} are the most accepted measures. However, this simple non-parametric analysis, depicted in Figure \ref{fge:14_decurto_and_dezarza} can serve as a unit test for GAN and other variational methods once the model is trained, and the authors have not seen it used at all in the community so far.

\setlength{\tabcolsep}{4pt}
\begin{table}[H]
\begin{center}
\caption{Interpretation of statistical measures given the proposed pipeline under study (Figure \ref{fge:14_decurto_and_dezarza}).\\}
\label{table:statistics}
\resizebox{0.7\columnwidth}{!}{%
\begin{tabular}{c|ccc}
\hline\noalign{\smallskip}
Test & Population & Result & Interpretation\\
\noalign{\smallskip}
\hline
\noalign{\smallskip}
1  & $C_{1}$ and $C_{2}$ & \begin{tabular}{@{}l@{}}
                                \cellcolor{green!24}\cmark \\
                                \xmark\\
                            \end{tabular} & \begin{tabular}{@{}l@{}}
                                \cellcolor{green!24}(a)\\
                                (b)\\
                            \end{tabular}\\ \hline
2 & $C_{2}$ &   \begin{tabular}{@{}l@{}}
                   \cmark \\
                   \cellcolor{green!24}\xmark\\
                \end{tabular} &
                \begin{tabular}{@{}l@{}}
                   (c)\\
                   \cellcolor{green!24}(d)\\
                \end{tabular}\\ \hline
3 & $C_{1}$ and $C_{2}$ &   \begin{tabular}{@{}l@{}}
                                \cellcolor{green!24}\cmark \\
                                \xmark\\
                            \end{tabular} &
                            \begin{tabular}{@{}l@{}}
                                \cellcolor{green!24}(e)\\
                                (f)\\
                            \end{tabular}
                            \\
\hline
\end{tabular}
}
\end{center}
\end{table}
\setlength{\tabcolsep}{1.4pt}
Description and interpretation of statistical measures are provided in Table~\ref{table:statistics}:
\begin{enumerate}[(a)]
\item Necessary condition but not sufficient to assert that both populations originate from the same distribution.
\item There is not enough statistical evidence to attest both populations samples originate from the same distribution.
\item With high probability the synthetic distribution generated is still close enough to the initial distribution of noise from the GAN architecture. The samples may not show enough fidelity, and there is probably bad generalization behavior.
\item The synthetic distribution is far from the initial distribution of noise and has deviated from the original Normal, and may be close to the target distribution.
\item If (a) then there is enough statistical evidence to confirm that both populations originate from the same distribution given this image descriptor. If (a) is not fulfilled, then we can only ascertain that the synthetic population is a good approximation.
\item There is not enough statistical evidence to attest both populations are from the same distribution.
\end{enumerate}

In Table \ref{table:statistical_measures} we can see evaluation test measures of homoscedasticity (T1), normality (T2) and goodness of fit (T3) on NASA Perseverance, AFHQ \cite{Choi2020} and MetFaces \cite{Karras2020}. According to the interpretation proposed in Table \ref{table:statistics}, we can conclude, given this image descriptor, that the models of Stylegan2-ada trained on AFHQ Cat and Wild are very good approximations of the original distributions (we accept the null hypothesis for goodness of fit), although we cannot conclude that the distributions are equal as the equality of variances is not assured. On AFHQ Dog, the model needs more training as T2 (normality of the synthetic distribution) is accepted, and therefore the learned distribution is close to the original white noise. The same conclusion holds true for the model trained on NASA Perseverance: more training is needed. For the case of MetFaces, the learned distribution is far from the original white noise, but we cannot accept the null hypothesis for goodness of fit (there are possible interpretations: there is overfit, the model needs more capacity to represent all the features from the original distribution, or it needs more training).

\setlength{\tabcolsep}{4pt}
\begin{table}[H]
\begin{center}
\caption{Evaluation of the statistical test measures of homoscedasticity (T1), normality (T2) and goodness of fit (T3) on AFHQ and MetFaces using state-of-the-art pretrained models of Stylegan2-ada \cite{Karras2020} and Stylegan3-ada \cite{Karras2021} and NASA, PERSEVERANCE.\\}
\label{table:statistical_measures}
\resizebox{\columnwidth}{!}{%
\begin{tabular}{c|cc|c|c|c}
\hline\noalign{\smallskip}
Model & \multicolumn{2}{c|}{Dataset} & T1 & T2 & T3 \\
\noalign{\smallskip}
\hline
\noalign{\smallskip}
\multirow{5}{*}{Stylegan2-ada} & NASA Perseverance &  & \xmark & \cmark & \xmark \\
\cline{2-6}
 & \multirow{3}{*}{AFHQ} & Cat  & \xmark & \cellcolor{green!24}\xmark & \cellcolor{green!24}\cmark \\
\cline{3-6}
&  & Dog  & \xmark & \cmark & \xmark  \\
\cline{3-6}
&  &Wild  & \xmark & \cellcolor{green!24}\xmark & \cellcolor{green!24}\cmark  \\
\cline{2-6}
& \multirow{3}{*}{MetFaces} & & \xmark & \cellcolor{green!24}\xmark & \xmark  \\
\cline{1-1}
$r$-Stylegan3-ada & & & \xmark & \cellcolor{green!24}\xmark & \xmark  \\
\cline{1-1}
$t$-Stylegan3-ada  & & & \xmark & \cellcolor{green!24}\xmark & \xmark  \\
\hline
\end{tabular}
}
\end{center}
\end{table}
\setlength{\tabcolsep}{1.4pt}

We have proposed statistical measures and a visualization pipeline to study and understand the data under consideration. However, the highly dimensional nature of images, and the fact that video streams are sequential, introduces a notion of time and space that our analysis has not taken into consideration. Indeed, the data consists on a sequence of images captured during a given lineal period of time following a specific path on the surface of Mars. To this effect, in the next section we borrow tools from harmonic analysis to provide further interpretation.

\section{Signature Transform and Harmonic Analysis}
\label{sn:signature}
The Signature Transform \cite{Bonnier2019,Kidger2021,Chevyrev2016,Liao2019,Morrill2021} is a roughly equivalent to Fourier; instead of extracting information about frequency, it extracts information about order and area. 
\\

Howbeit, the Signature Transform differs from Fourier by the fact that it utilizes a basis of the space of functions of paths, a more general case to the basis of space of paths found in the preceding. 
\\

Following \cite{Bonnier2019}, the truncated signature of order $N$ of the path $\mathbf x$ is defined as a collection of coordinate iterated integrals
\begin{align}
& \medmath{\mathrm{S}^{N}(\mathbf x) =} \nonumber \\ & \medmath{\left(\left( \underset{0 < t_1 < \cdots < t_a < 1}{\int\cdots\int} \prod_{c = 1}^a \frac{\mathrm d f_{z_c}}{\mathrm dt}(t_c) \mathrm dt_1 \cdots \mathrm dt_a \right)_{\!\!1 \le z_1, \ldots, z_a \le d}\right)_{\!\!1\le a \le N}.}
\end{align}

The Signature is a homomorphism from the monoid of paths into the grouplike elements of a closed tensor algebra, see Equation \ref{en:01_decurto_and_dezarza}. It provides a graduated summary of the path $\mathbf x$. These extracted features of a path are at the center of the definition of a rough path \cite{Lyons2014}; they remove the necessity to take into account the inner detailed structure of the path.

\begin{align}
\mathrm{S}:\left\{f \in F \ | \ f:[x,y]\to E 
= \mathbb{R}^{d}\right\} \longrightarrow T(E) \\ \textnormal{where } T(E) = T(\mathbb{R}^{d}) =  \prod_{c=0}^{\infty} \left ( \mathbb{R}^{d}\right)^{\otimes c}.
\label{en:01_decurto_and_dezarza}
\end{align}

It has many advantages over other tools of harmonic analysis for ML. It is a universal non-linearity, which means that every continuous function of the input stream may be approximated arbitrary well by a linear function of its signature. Also, among other properties, it presents outstanding robustness behavior to missing or irregularly sampled data, along with optional invariance in terms of translation and sampling. It has recently been introduced in the context of Deep Learning to add some structure to the learning process, and it seems a promising tool in Generative Models and Reinforcement Learning, as well as a good theoretical framework. It mainly works on streams of data which could describe from video sequences to our entire life experiences. Scilicet, under the correct assumptions and the right application, it could potentially compress all human experiences in a representation that could be stored and processed efficiently. Here we propose to do a preliminary study in terms of harmonic analysis, and understand its properties to compare the original and synthetic samples.

\begin{figure}[ht]
\centering
\subfloat[]{\includegraphics[scale=0.3]{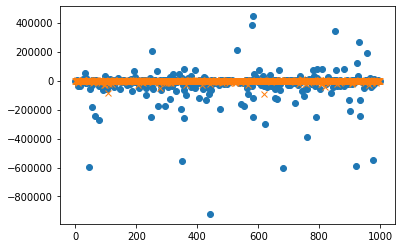}}
\subfloat[]{\includegraphics[scale=0.3]{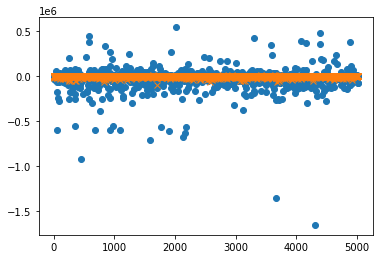}}\\
\subfloat[]{\includegraphics[scale=0.3]{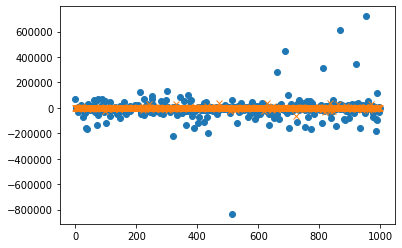}}
\subfloat[]{\includegraphics[scale=0.3]{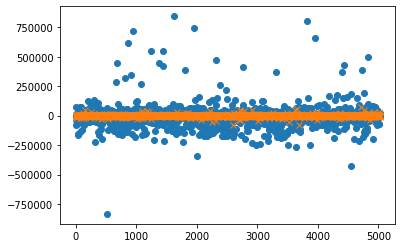}}
\caption{Spectrum of the mean signature (a,b) and log-signature (c,d) of a $64$-dimensional path up to level $3$ of original (`o') against $1000$ (a,c) and $5000$ (b,d) synthetic (`x') samples.}
\label{fge:07_decurto_and_dezarza}
\end{figure}

The Signature \cite{Lyons2014,Kiraly2019,Graham2013,Chang2019,Fermanian21} of an input stream of data encodes the order in which data arrives without caring precisely when it arrives. This property, which is known as invariance to time reparameterizations \cite{Lyons1998}, makes it an ideal candidate to measure GAN generated distributions against an original data stream. That is to say, when sampling the GAN model, instances of the latent space are retrieved in no specific order, although the original data is by definition time dependent, as recorded video streams or image captured by sensors are constrained and bounded by time physics. However, GANs are not able to generate yet data lineally in time and space, and thus the comparison using other methods may be biased, or not take into account all the relevant cues.
\\

Withal, it is important to note that the number of components of the truncated signature does not depend on the number of data samples into consideration. Namely, it maps the infinite-dimensional space of streams of data $\mathcal{S}(\mathbb{R}^{d})$ into a finite-dimensional space of dimension $(d^{N+1}-1)/(d-1)$, where $N$ corresponds to the order of the truncated signature, which makes it very appropriate to process long sequential data with varying length or unevenly sampled data.
\\

At the same time, we can introduce the concept of log-signature \cite{Liao2019,Morrill2021}, which is a more compact representation than the Signature.

\begin{definition}
If $\gamma _{t}\in E$ is a path segment and $S$ is its Signature then 
\begin{eqnarray*}
S &=&1+S^{1}+S^{2}+\ldots \ \forall c,\ S^{c}\in E^{\otimes c} \\
\log \left( 1+x\right) &=&x-x^{2}/2+\ldots \\
\log S &=&\left( S^{1}+S^{2}+\ldots \right) -\left( S^{1}+S^{2}+\ldots
\right) ^{2}/2+\ldots
\end{eqnarray*}%
The series $\log S=\left( S^{1}+S^{2}+\ldots \right) -\left(
S^{1}+S^{2}+\ldots \right) ^{2}/2+\ldots $ which is well defined, is
referred to as the log-signature of $\gamma .$
\end{definition}

Unlike the Signature, the log-signature does not guarantee universality \cite{Lyons2014}, and thus it needs to be combined with non-linear models for learning. However, it is empirically more robust to sparsely sampled data. There is a one-to-one correspondence between the Signature and the log-signature as the logarithm map is bijective \cite{Lyons2007,Liao2019}. This statement also holds true for the truncated case up to the same degree.
\\

In this line of work, we perform a comparison of the mean signature and log-signature of original against synthetic samples at size $64\times64$, and observe that synthetic samples encompass the most relevant information from the original harmonic distribution, see Figure \ref{fge:07_decurto_and_dezarza}. We compare against a set of $1000$ and $5000$ synthetic samples, and each instance is considered to be a path $\mathbf x$ of dimension $64$ to which we apply the Signature and log-signature transforms.

\begin{figure*}[ht]
\centering
\includegraphics[scale=0.47]{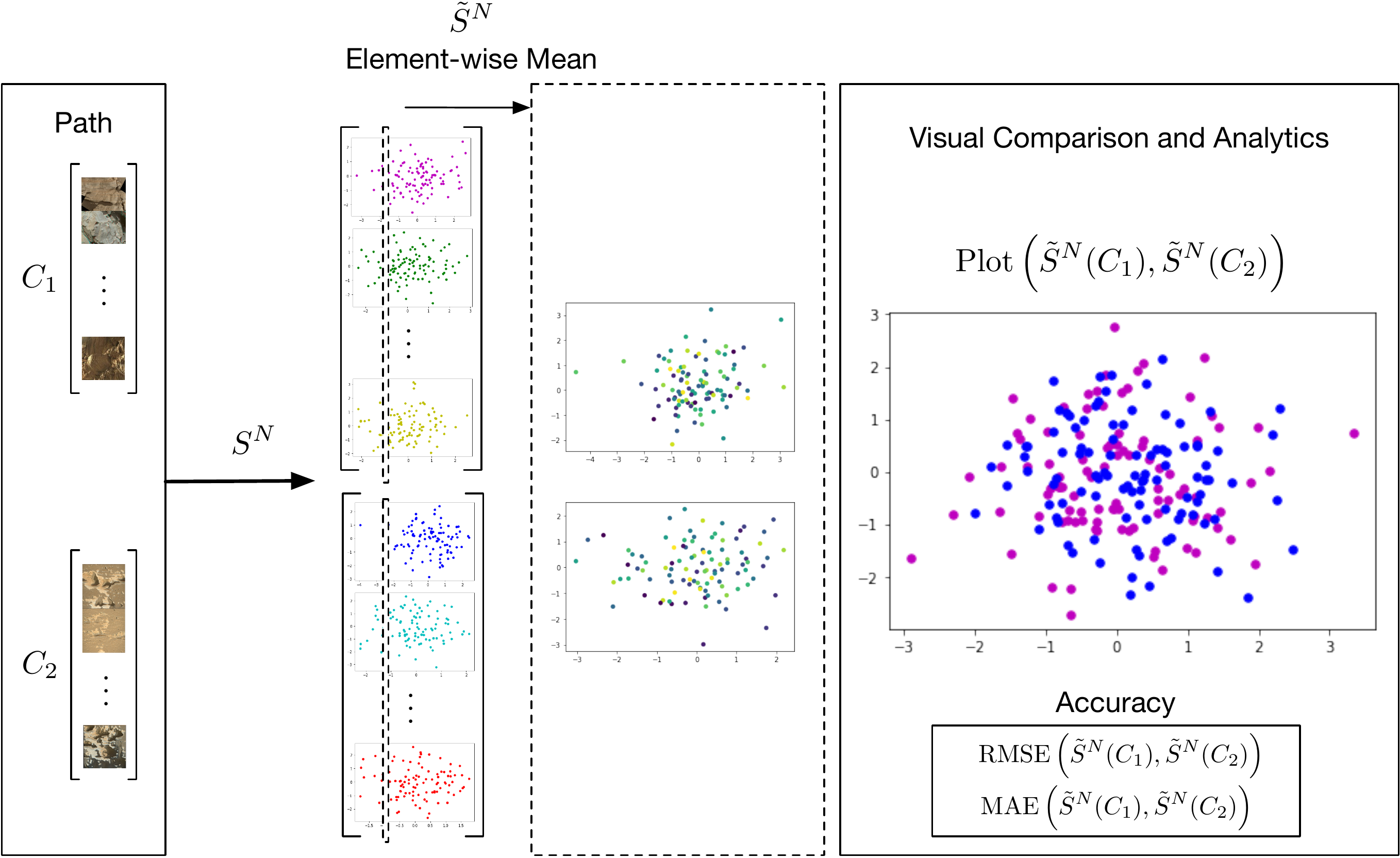}
\caption{Visual explanation of the use of $\tilde{S}^{N}$ to analyze GAN convergence. Samples are resized at $64\times64$ and transformed to grayscale previous to the computation of the signatures. The procedure used for log-signature $\log \tilde{S}^{N}$ is analogous.}
\label{fge:21_decurto_and_dezarza}
\end{figure*}

\subsection{RMSE and MAE Signature and Log-signature}
\begin{figure*}[ht]
\centering
\subfloat[]{\includegraphics[scale=0.3]{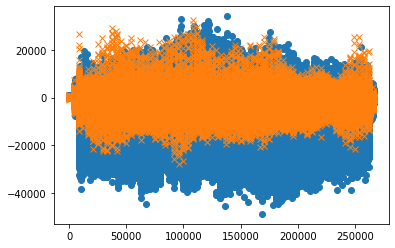}}
\subfloat[]{\includegraphics[scale=0.3]{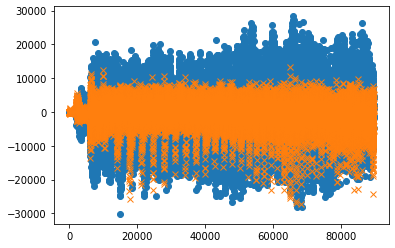}}
\caption{Spectrum of the element-wise mean of the signatures (left) and log-signatures (right) of order 3 and size $64\times64$ of original (`o') against synthetic (`x') samples.}
\label{fge:19_decurto_and_dezarza}
\end{figure*}

We propose to use the element-wise mean of the truncated signatures $\tilde{S}^{N}$, illustrated in Figure \ref{fge:21_decurto_and_dezarza}, to analyse the convergence of GAN learned models by the use of RMSE (Root Mean Squared Error) and MAE (Mean Absolute Error); we name the measures RMSE and MAE Signature, and RMSE and MAE log-signature. For instance, in Figure \ref{fge:19_decurto_and_dezarza} we can observe that the model is attaining good convergence, although is not capturing all the information present in the original distribution.
\\

We can understand RMSE and MAE through the element-wise mean as a score function on top of the Signature Transform that is able to measure the quality of the generated distribution. This way of understanding these measures will be important for future uses, allowing us to generalize them on other applications, or possibly other transforms \cite{DeCurto22_3}. 

RMSE and MAE Signature and log-signature can be used not only to compare models, but also to keep performance of training across several epochs, and analytically detect overfitting, as highlighted in Table \ref{table:rmse}. Although all measures capture information about the visual cues present in the distributions, RMSE and MAE Signature and MAE log-signature are more accurate at keeping track of the GAN training procedure convergence, while RMSE log-signature is less precise.

\setlength{\tabcolsep}{4pt}
\begin{table}[H]
\begin{center}
\caption{RMSE and MAE, SIGNATURE and LOG-SIGNATURE across several iterations of training of Stylegan2-ada (lower is better). Our synthetic samples are generated using the model $798$ which achieves the highest accuracy on RMSE and MAE, SIGNATURE and LOG-SIGNATURE.\\}
\label{table:rmse}
\resizebox{\columnwidth}{!}{%
\begin{tabular}{l|ccccc}
\hline\noalign{\smallskip}
Iteration Stylegan2-ada & 193 & 371 & 596 & 798 & 983\\
\noalign{\smallskip}
\hline
\noalign{\smallskip}
RMSE Signature  & 15617 & 13336 & 12353 & \cellcolor{blue!24}\textbf{11601} & 25699\\
MAE Signature  & 11072 & 10686 & 9801 & \cellcolor{blue!24}\textbf{9086} & 19481\\
RMSE log-signature  & 9882 & 7563 & \cellcolor{blue!24}\textbf{7354} & 7397 & 15621\\
MAE log-signature  & 6467 & 5955 & 5724 & \cellcolor{blue!24}\textbf{5717} & 12063\\
\hline
\end{tabular}
}
\end{center}
\end{table}
\setlength{\tabcolsep}{1.4pt}

To further expand the concepts illustrated in this section we analytically describe the abstraction of a set of images as a unevenly sampled stream of data, e.g. a path, as well as the definitions to measure the similarity between image distributions.
\\

We can understand a stream of data $\mathbf{x}\in \mathcal{S}(\mathbb{R}^{d})$ as a discrete representation of a path. 

\begin{definition}\label{def:s}
	Let $\mathbf x=(x_1, \ldots, x_n)\in \mathcal S(\mathbb R^d)$ be a stream of data. Let $X$ be a linear interpolation of $\mathbf x$. Then the signature of $\mathbf x$ is defined as
	\begin{equation}
	\mathrm{S}(\mathbf x) = \mathrm{S}(X)
	\end{equation}
	and the truncated signature of order $N$ of $\mathbf x$ is defined as
	\begin{equation}
	\mathrm{S}^N(\mathbf x) = \mathrm{S}^N(X).
	\end{equation}
\end{definition}

This definition of the signature of a stream of data is independent of the choice of linear interpolation of $X$ by the invariance to time reparameterizations \cite{Bonnier2019}.
\\

\begin{definition}
    Given a set of truncated signatures of order $N$, $\left\{ \mathrm{S}^{N}_{c}(\mathbf{x}_{c})\right\}_{c=1}^{m}$, the element-wise mean is defined by
    \begin{equation}
    \tilde{\mathrm{S}}^{N}(x^{(z)})=\frac{1}{m} \sum_{c=1}^{m}\mathrm{S}^{N}_{c}(x^{(z)}_{c}),    
	\end{equation}
	where $z \in \{1,\ldots,n\}$ is the specific component index of the given signature.
\end{definition}

Then RMSE and MAE Signature, whose results are presented in Tables \ref{table:rmse} and \ref{table:evaluation}, can be defined as follows.

\begin{definition}
    Given $n$ components of the element-wise mean of the signatures $\{y^{(c)}\}^{n}_{c=1}\subseteq T(\mathbb{R}^{d})$ from the model chosen as a source of synthetic samples, and the same number of components of the element-wise mean of the signatures $\{x^{(c)}\}^{n}_{c=1}\subseteq T(\mathbb{R}^{d})$ from the original distribution, then we define the Root Mean Squared Error (RMSE) and Mean Absolute Error (MAE) by
    \begin{equation}
    \medmath{\textnormal{RMSE}\left(\left\{x^{(c)}\right\}^{n}_{c=1},\left\{y^{(c)}\right\}^{n}_{c=1}\right) = \sqrt{\frac{1}{n} \sum_{c=1}^{n} \left( y^{(c)} - x^{(c)}\right)^{2}},}
	\end{equation}
	and
    \begin{equation}
    \medmath{\textnormal{MAE}\left(\left\{x^{(c)}\right\}^{n}_{c=1},\left\{y^{(c)}\right\}^{n}_{c=1}\right) = \frac{1}{n} \sum_{c=1}^{n} | y^{(c)} - x^{(c)} |.}
	\end{equation}
\end{definition}
The case for log-signature is analogous.

\subsection{Evaluation}
\begin{figure}[ht]
\centering
\subfloat[]{\includegraphics[scale=0.2]{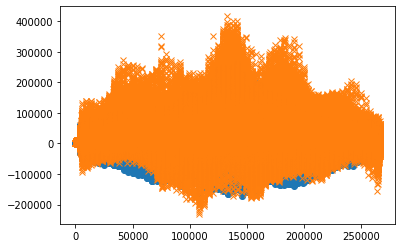}}
\subfloat[]{\includegraphics[scale=0.2]{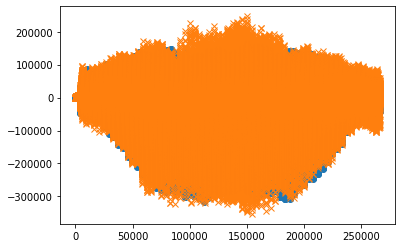}}
\subfloat[]{\includegraphics[scale=0.2]{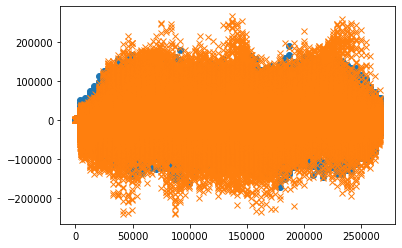}} \\
\subfloat[]{\includegraphics[scale=0.2]{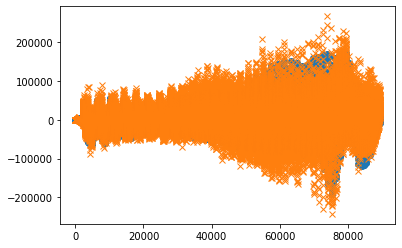}}
\subfloat[]{\includegraphics[scale=0.2]{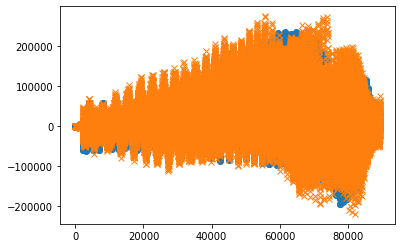}}
\subfloat[]{\includegraphics[scale=0.2]{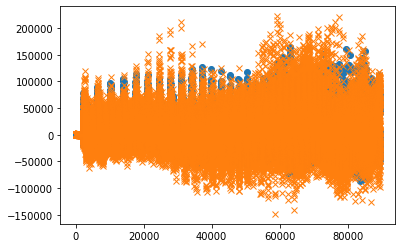}}
\caption{Spectrum of the element-wise mean of the signatures $\tilde{S}^{3}$ (top) and log-signatures $\log \tilde{S}^{3}$ (bottom) of order 3 and size $64\times64$ of original (`o') against synthetic (`x') samples. (a,d): AFHQcat, (b,e): AFHQdog, (c,f): AFHQwild.}
\label{fge:23_decurto_and_dezarza}
\end{figure}

\begin{figure}[ht]
\centering
\subfloat[]{\includegraphics[scale=0.2]{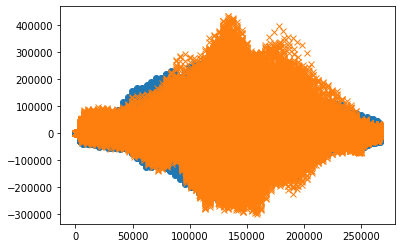}}
\subfloat[]{\includegraphics[scale=0.2]{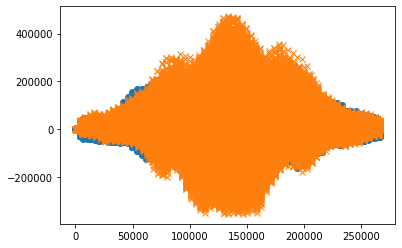}}
\subfloat[]{\includegraphics[scale=0.2]{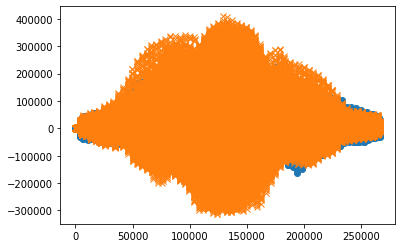}}\\
\subfloat[]{\includegraphics[scale=0.2]{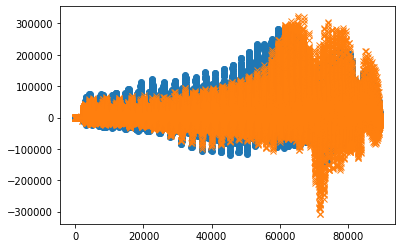}}
\subfloat[]{\includegraphics[scale=0.2]{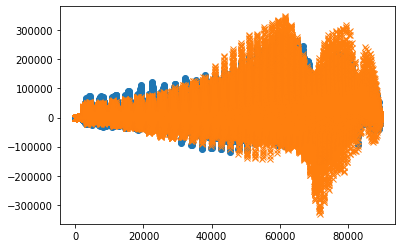}}
\subfloat[]{\includegraphics[scale=0.2]{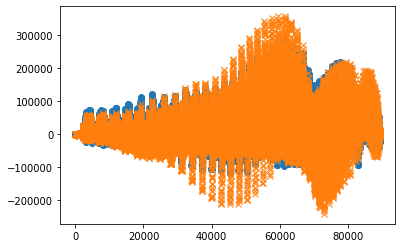}}
\caption{Spectrum comparison of the element-wise mean of the signatures $\tilde{S}^{3}$ (top) and log-signatures $\log \tilde{S}^{3}$ (bottom) of order 3 and size $64\times64$ of original (`o') against synthetic (`x') samples from MetFaces. (a,d): Stylegan2-ada, (b,e): $r$-Stylegan3-ada, (c,f): $t$-Stylegan3-ada.}
\label{fge:27_decurto_and_dezarza}
\end{figure}

We enclose results of the proposed measures using several state-of-the-art pretrained models; these results are presented in Table \ref{table:evaluation}. For the evaluation and testing we use the standard AFHQ dataset \cite{Choi2020} classes `cat', `dog' and `wild', along with MetFaces \cite{Karras2020}, together with the corresponding pretrained models. To compute RMSE and MAE $\tilde{S}^{N}$ and $\log \tilde{S}^{N}$ we generate $1000$ synthetic samples of each model and compare against the full original dataset. The samples are transformed to grayscale and resized at $64\times64$ previous to the Signature Transform. Visual comparison of the spectrum is provided in Figures \ref{fge:23_decurto_and_dezarza} and \ref{fge:27_decurto_and_dezarza}, where it can be seen that the trained models are actually learning the empirical distribution of the original data.\\

In Table \ref{table:evaluation} we compare the recently developed models $\{r,t\}$-Stylegan3-ada \cite{Karras2021} against Stylegan2-ada using MetFaces, where we can see that $t$-Stylegan3-ada clearly outperforms Stylegan2-ada and $r$-Stylegan3-ada, which agree with the FID results reported in \cite{Karras2021}, as shown in Table \ref{table:fid} where we can observe that FID very closely resembles the behavior of RMSE $\tilde{S}^{3}$. Even so, our metrics are effective and efficient. Visual comparison of the spectrum of the Signatures for the given dataset can be seen in Figure \ref{fge:27_decurto_and_dezarza}. Computation is done at the CPU in seconds, orders of magnitude faster and with fewer resources than FID or MS-SSIM.

\setlength{\tabcolsep}{4pt}
\begin{table}[H]
\begin{center}
\caption{RMSE and MAE, SIGNATURE and LOG-SIGNATURE evaluation and comparison on AFHQ and MetFaces using state-of-the-art pretrained models of Stylegan2-ada \cite{Karras2020} and Stylegan3-ada \cite{Karras2021}.\\}
\label{table:evaluation}
\resizebox{\columnwidth}{!}{%
\begin{tabular}{c|lc|c|c|c|c}
\hline\noalign{\smallskip}
Model & \multicolumn{2}{c|}{Dataset} & RMSE $\tilde{S}^{3}$ & MAE $\tilde{S}^{3}$ & RMSE $\log \tilde{S}^{3}$ & MAE $\log \tilde{S}^{3}$\\
\noalign{\smallskip}
\hline
\noalign{\smallskip}
\multirow{4}{*}{Stylegan2-ada} & \multirow{3}{*}{AFHQ} & Cat  & 61450 & 45968 & 29201 & 22297 \\
\cline{3-7}
&  & Dog  & 38861 & 30441 & 31686 & 24612 \\
\cline{3-7}
&  &Wild  & 33306 & 25578 & 26622 & 20359 \\
\cline{2-7}
& \multirow{3}{*}{MetFaces} & & 33247 & 23428 & 25685 & 18071 \\
\cline{1-1}
$r$-Stylegan3-ada & & & 34977 & 22799 & 24707 & 16539 \\
\cline{1-1}
\rowcolor{blue!24}
$t$-Stylegan3-ada  & & & \textbf{30894} & \textbf{19872} & \textbf{21560} & \textbf{13761} \\
\hline
\end{tabular}
}
\end{center}
\end{table}
\setlength{\tabcolsep}{1.4pt}

\setlength{\tabcolsep}{4pt}
\begin{table}[H]
\begin{center}
\caption{Evaluation and comparison of FID (as reported in \cite{Karras2021}) and RMSE $\tilde{S}^{3}$ on MetFaces.\\}
\label{table:fid}
\resizebox{0.52\columnwidth}{!}{%
\begin{tabular}{c|c|c}
\hline\noalign{\smallskip}
Model & FID & RMSE $\tilde{S}^{3}$\\
\noalign{\smallskip}
\hline
Stylegan2-ada & 15.22 & 33247\\
\cline{1-1}
$r$-Stylegan3-ada & 15.33 & 34977\\
\cline{1-1}
$t$-Stylegan3-ada  & \cellcolor{blue!24}\textbf{15.11} & \cellcolor{blue!24}\textbf{30894}  \\
\hline
\end{tabular}
}
\end{center}
\end{table}
\setlength{\tabcolsep}{1.4pt}

\section{Exploration of the samples}
\label{sn:exploration}

\begin{figure*}[ht]
\centering
\subfloat[]{\includegraphics[scale=0.21]{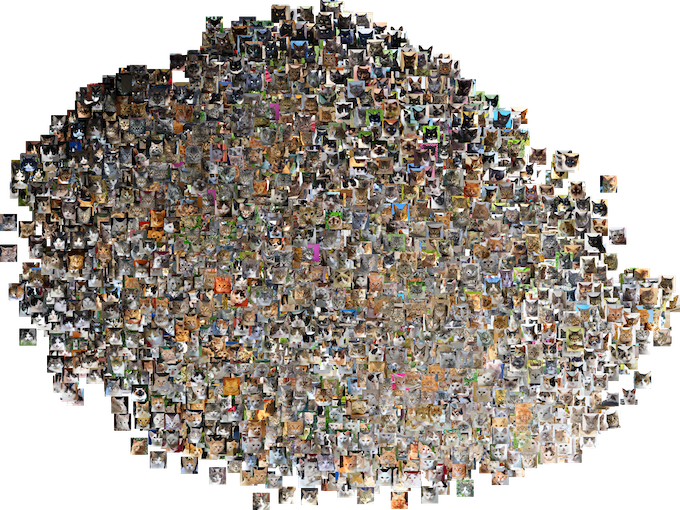}}
\subfloat[]{\includegraphics[scale=0.21]{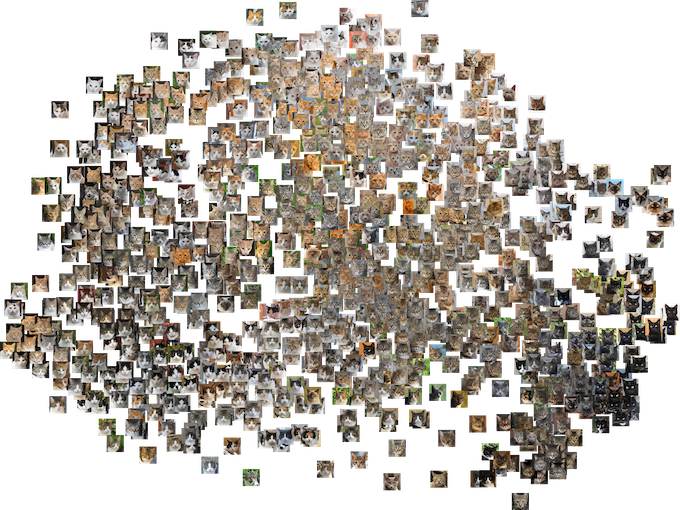}}\\
\subfloat[]{\includegraphics[scale=0.21]{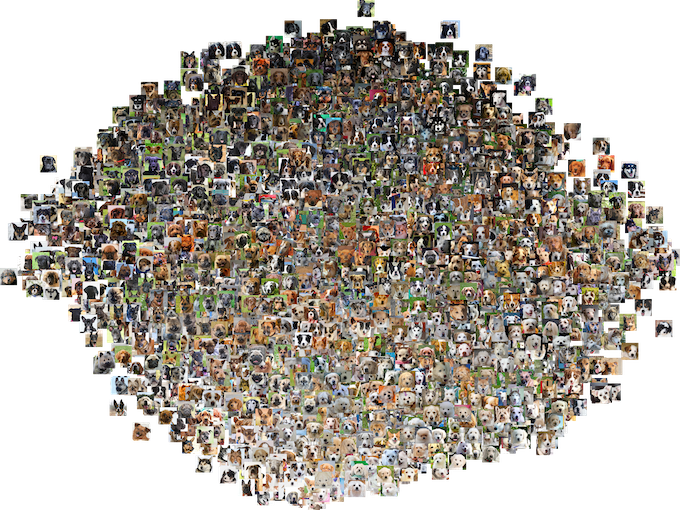}}
\subfloat[]{\includegraphics[scale=0.21]{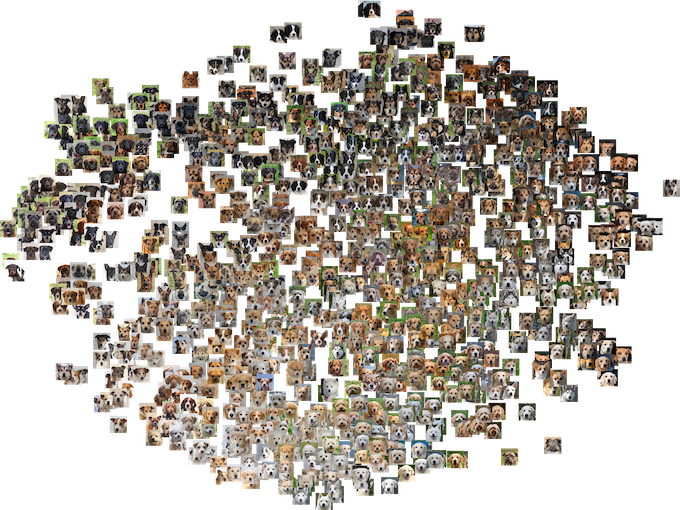}}\\
\subfloat[]{\includegraphics[scale=0.21]{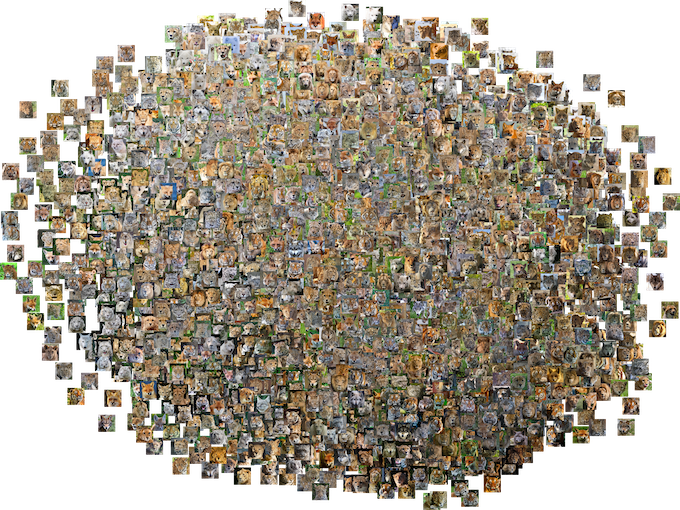}}
\subfloat[]{\includegraphics[scale=0.21]{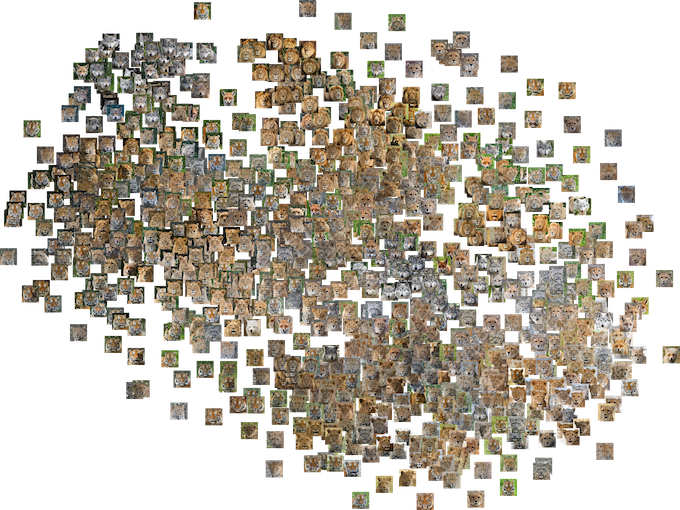}}
\caption{Visualization of PCA Adaptive t-SNE on original (left) versus synthetic (right) samples of AFHQ Cat (a,b), Dog (c,d) and Wild (e,f) using Stylegan2-ada.}
\label{fge:b_decurto_and_dezarza}
\end{figure*}

\begin{figure}[ht]
\centering
\subfloat[]{\includegraphics[scale=0.21]{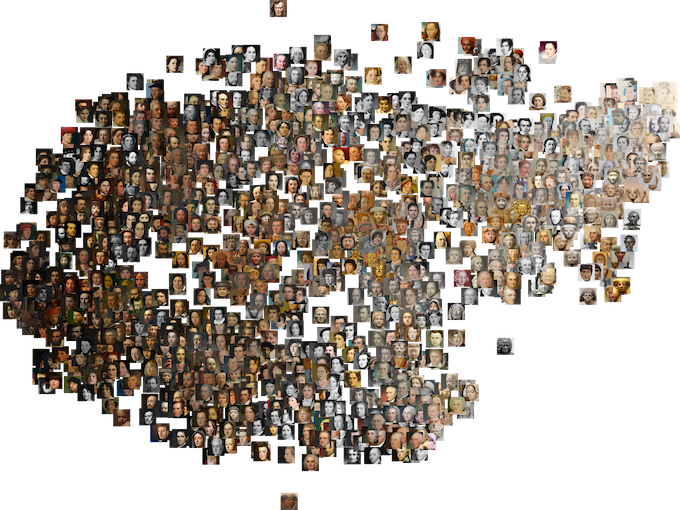}}\\
\subfloat[]{\includegraphics[scale=0.11]{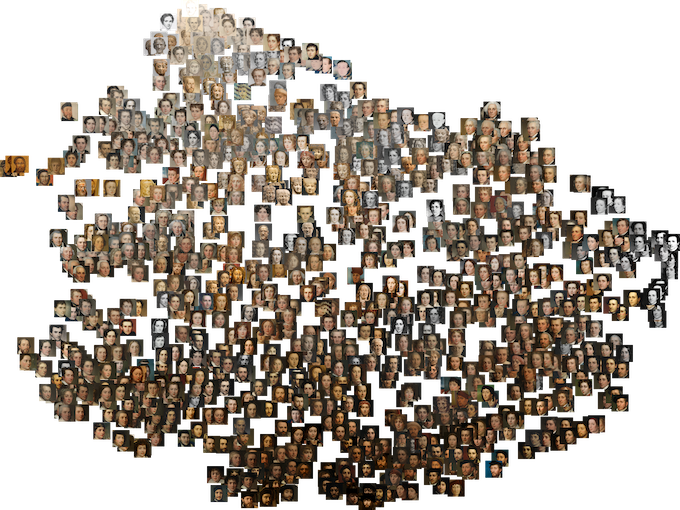}}
\subfloat[]{\includegraphics[scale=0.11]{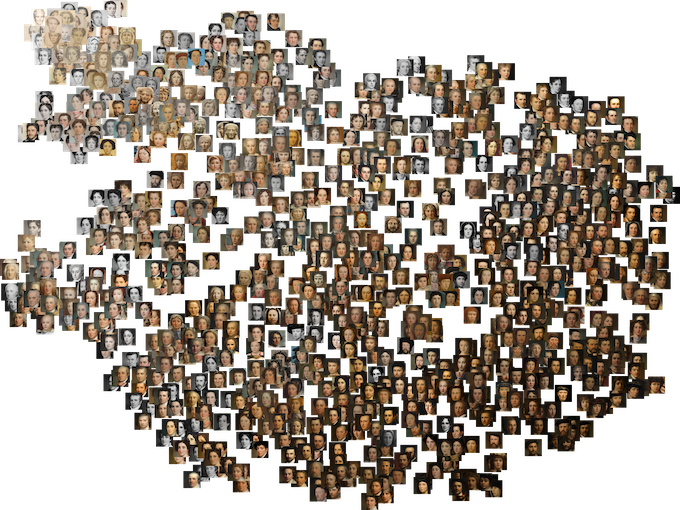}}
\subfloat[]{\includegraphics[scale=0.11]{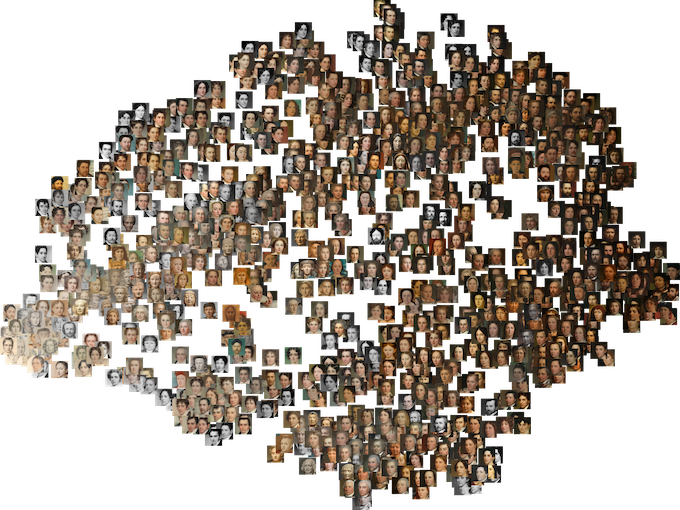}}
\caption{Visualization of PCA Adaptive t-SNE on original (top) versus synthetic (bottom) samples of MetFaces using Stylegan2-ada (b), $r$-Stylegan3-ada (c) and $t$-Stylegan3-ada (d).}
\label{fge:b_2_decurto_and_dezarza}
\end{figure}

\begin{figure}[ht]
\centering
\subfloat[]{\includegraphics[scale=0.21]{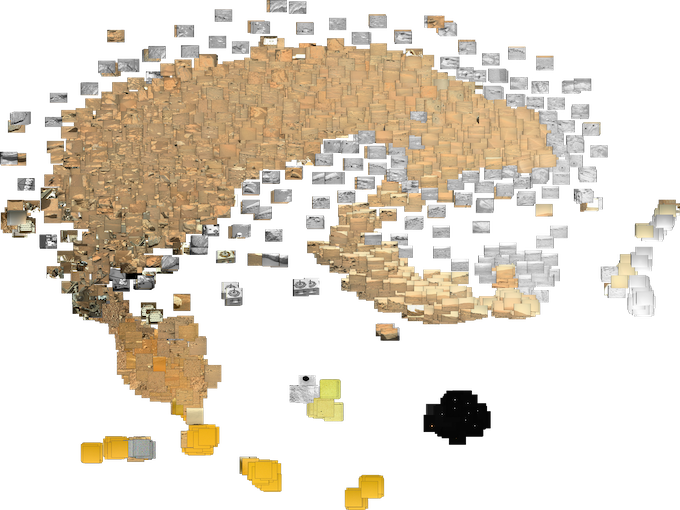}}\\
\subfloat[]{\includegraphics[scale=0.12]{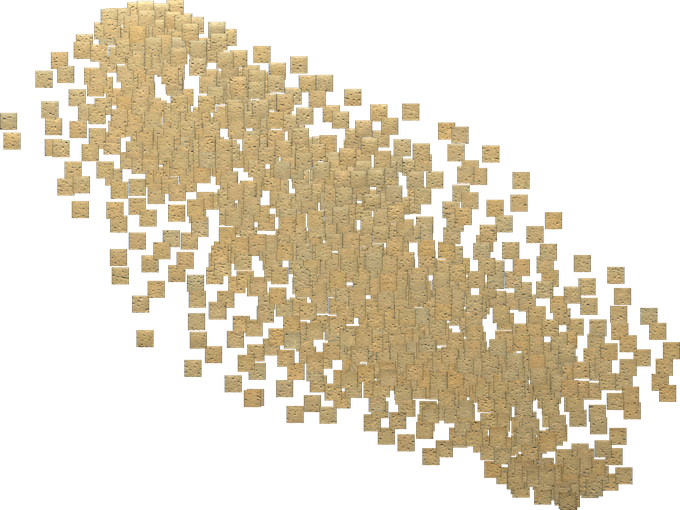}}
\subfloat[]{\includegraphics[scale=0.12]{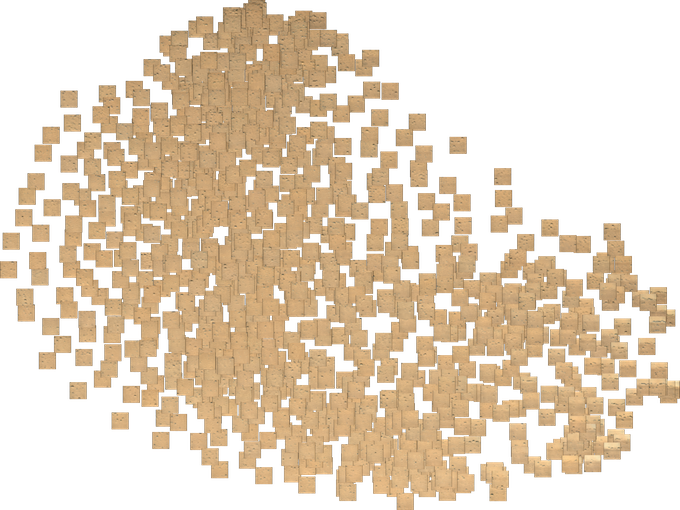}}
\subfloat[]{\includegraphics[scale=0.12]{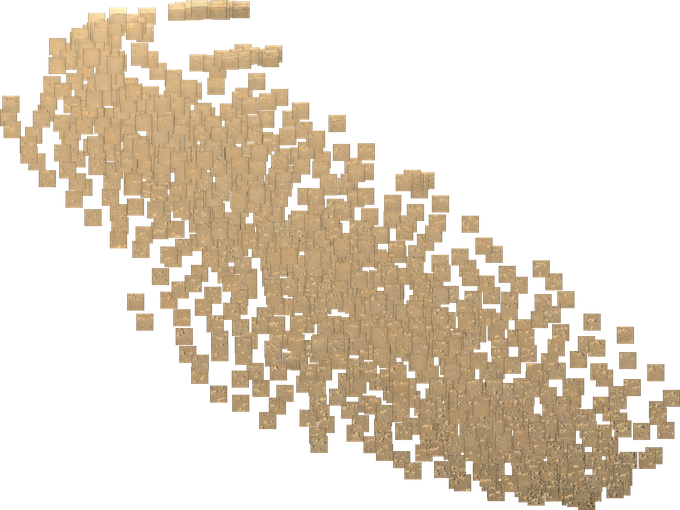}}\\
\subfloat[]{\includegraphics[scale=0.13]{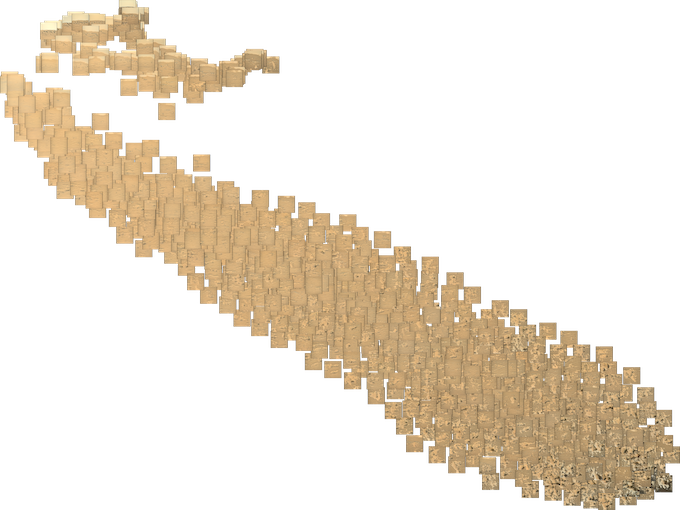}}
\subfloat[]{\includegraphics[scale=0.13]{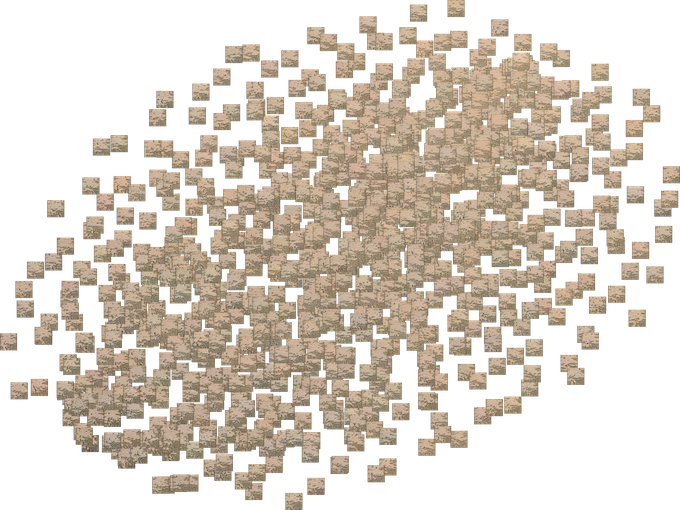}}
\caption{Visualization of PCA Adaptive t-SNE on original (top) versus synthetic (bottom) samples of NASA Perseverance using Stylegan2-ada across several epoch iterations: $193$ (b), $371$ (c), $596$ (d), $798$ (e) and $983$ (d).}
\label{fge:b_3_decurto_and_dezarza}
\end{figure}

In Figure \ref{fge:b_decurto_and_dezarza} and \ref{fge:b_2_decurto_and_dezarza} we visualize using PCA Adaptive t-SNE the sets of images of AFHQ and MetFaces, original and synthetic, used in the evaluations in Tables \ref{table:evaluation} and \ref{table:statistical_measures}. We can observe, for instance, that the synthetic samples of AFHQ Cat and Wild resemble very much the original distribution, both in variability and quality, while AFHQ Dog lacks some variability but achieves very good quality samples, which agrees with the analytical interpretation of the proposed statistical measures, shown in Table \ref{table:statistical_measures}. Visualization across several epochs of training of NASA Perseverance can be seen on Figure \ref{fge:b_3_decurto_and_dezarza}.

In Figure \ref{fge:b_2_decurto_and_dezarza} we can perceive that the synthetic samples generated with $t$-Stylegan3-ada show better quality than Stylegan2-ada and $r$-Stylegan3-ada, and the model is clearly learning the original distribution. However, there is room for improvement in terms of variability and scope. These arguments agree with RMSE and MAE Signature and log-signature, as shown in Table \ref{table:evaluation}. 

In Figure \ref{fge:b_3_decurto_and_dezarza} we can see the PCA Adaptive t-SNE visualization of original and synthetic samples from NASA Perseverance across several epochs of training following the performance results presented in Table \ref{table:rmse}. Results shown in Figure \ref{fge:03_decurto_and_dezarza} and Table \ref{table:statistical_measures} originate from epoch iteration $798$, which achieves the best RMSE Signature and MAE Signature and log-signature. Visual inspection of iteration $798$ corroborates the interpretation from Table \ref{table:statistical_measures} that the samples are close to the original WHITE noise, and that the model is far from being a good representation of the original distribution as it only captures a small subset of the visual cues present in the data. Furthermore, RMSE and MAE Signature and log-signature in Table \ref{table:rmse} correctly detect overfitting on iteration $983$.

\section{Conclusions}
\label{sn:conclusions}

GAN evaluation has been one of the central research efforts of the community of computer vision during these last years. The ability of the networks to generate high-fidelity samples has inspired researchers all over the world to work on the topic. However, although many variants of the original successful DCGAN architecture are able to generate very realistic samples, neither the advance in proposing metrics to assess the imagery has been effectual, nor the ability of the metrics to guarantee some level of robustness, and overall description of the resultant distribution. The best effort of them being FID suffers from high-computation time and use of GPU resources; it depends mainly on an inception module that extracts features from lots of samples rather than from analytical measures that quantify properly their characteristics.  
\\

We are the first to propose the use of the Signature Transform to assess GAN convergence by introducing RMSE and MAE Signature and log-signature. The measures are reliable, consistent, efficient and easy to compute. Additionally, an effective methodology to test the goodness-of-fit according to the original distribution by the use of simple statistical methods is also proposed, being the first to be able to reduce the amount of computation for accurate GAN Synthetic image quality assessment to the order of seconds. Worth to mention is the proposal of a taxonomical pipeline to systematically assess the resultant distributions using a non-parametric test. Lastly, we also introduce an adaptive technique based on t-SNE and PCA that, without the need of hyperparameter tuning, puts forward exceptional visualization capabilities.
\\

Future work that could be pursued under these assumptions, among others, is to increase the complexity of the descriptor, extend the proposed score functions on top of the Signature Transform to be used in other tasks or use the metrics inside the training loop to assess convergence and help the networks train faster.

\bibliographystyle{IEEEtran}


\begin{thebibliography}{10}
\providecommand{\url}[1]{#1}
\csname url@samestyle\endcsname
\providecommand{\newblock}{\relax}
\providecommand{\bibinfo}[2]{#2}
\providecommand{\BIBentrySTDinterwordspacing}{\spaceskip=0pt\relax}
\providecommand{\BIBentryALTinterwordstretchfactor}{4}
\providecommand{\BIBentryALTinterwordspacing}{\spaceskip=\fontdimen2\font plus
\BIBentryALTinterwordstretchfactor\fontdimen3\font minus
  \fontdimen4\font\relax}
\providecommand{\BIBforeignlanguage}[2]{{%
\expandafter\ifx\csname l@#1\endcsname\relax
\typeout{** WARNING: IEEEtran.bst: No hyphenation pattern has been}%
\typeout{** loaded for the language `#1'. Using the pattern for}%
\typeout{** the default language instead.}%
\else
\language=\csname l@#1\endcsname
\fi
#2}}
\providecommand{\BIBdecl}{\relax}
\BIBdecl

\bibitem{Goodfellow14}
I.~Goodfellow, J.~Pouget-Abadie, M.~Mirza, B.~Xu, D.~Warde-Farley, S.~Ozair,
  A.~Courville, and Y.~Bengio, ``Generative adversarial networks,''
  \emph{NIPS}, vol.~27, 2014.

\bibitem{Heusel17}
M.~Heusel, H.~Ramsauer, T.~Unterthiner, B.~Nessler, and S.~Hochreiter, ``{GANs}
  trained by a two time-scale update rule converge to a local nash
  equilibrium,'' \emph{NIPS}, vol.~30, 2017.

\bibitem{Lyons2014}
T.~Lyons, ``Rough paths, signatures and the modelling of functions on
  streams,'' \emph{Proceedings of the International Congress of
  Mathematicians}, 2014.

\bibitem{Girshick14}
R.~Girshick, J.~Donahue, T.~Darrell, and J.~Malik, ``Rich feature hierarchies
  for accurate object detection and semantic segmentation,'' \emph{IEEE
  Conference on Computer Vision and Pattern Recognition}, 2014.

\bibitem{He15}
K.~He, X.~Zhang, S.~Ren, and J.~Sun, ``Delving deep into rectifiers: Surpassing
  human-level performance on {ImageNet} classification,'' \emph{IEEE
  International Conference on Computer Vision}, 2015.

\bibitem{Girshick15}
R.~Girshick, ``Fast r-cnn,'' \emph{{IEEE} International Conference on Computer
  Vision}, 2015.

\bibitem{Long15}
J.~Long, E.~Shelhamer, and T.~Darrell, ``Fully convolutional networks for
  semantic segmentation,'' \emph{IEEE Conference on Computer Vision and Pattern
  Recognition}, 2015.

\bibitem{Redmon16}
J.~Redmon, S.~Divvala, R.~Girshick, and A.~Farhadi, ``You only look once:
  Unified, real-time object detection,'' \emph{{IEEE} Conference on Computer
  Vision and Pattern Recognition}, 2016.

\bibitem{Gatys16}
L.~A. Gatys, A.~S. Ecker, and M.~Bethge, ``Image style transfer using
  convolutional neural networks,'' \emph{{IEEE} International Conference on
  Computer Vision}, 2016.

\bibitem{Gatys16_2}
L.~A. Gatys, M.~Bethge, A.~Hertzmann, and E.~Shechtman, ``Preserving color in
  neural artistic style transfer,'' \emph{arXiv:1606.05897}, 2016.

\bibitem{He17}
K.~He, G.~Gkioxari, P.~Doll\'ar, and R.~Girshick, ``Mask r-cnn,'' \emph{{IEEE}
  International Conference on Computer Vision}, 2017.

\bibitem{Vaswani2017}
A.~Vaswani, N.~Shazeer, N.~Parmar, J.~Uszkoreit, L.~Jones, A.~N. Gomez,
  L.~Kaiser, and I.~Polosukhin, ``Attention is all you need,'' \emph{NIPS},
  2017.

\bibitem{Wang18_2}
T.~Wang, M.~Liu, J.~Zhu, G.~Liu, A.~Tao, J.~Kautz, and B.~Catanzaro,
  ``Video-to-video synthesis,'' \emph{NIPS}, 2018.

\bibitem{Karras18}
T.~Karras, T.~Aila, S.~Laine, and J.~Lehtinen, ``Progressive growing of {GANs}
  for improved quality, stability, and variation,'' \emph{ICLR}, 2018.

\bibitem{DeZarza22}
I.~{de Zarzà}, J.~{de Curtò}, and C.~T. Calafate, ``Detection of glaucoma
  using three-stage training with {EfficientNet},'' \emph{Intelligent Systems
  with Applications}, vol.~16, p. 200140, 2022.

\bibitem{Chen17}
L.~Chen, G.~Papandreou, I.~Kokkinos, K.~Murphy, and A.~L. Yuille, ``{DeepLab:
  Semantic Image Segmentation with Deep Convolutional Nets, Atrous Convolution,
  and Fully Connected CRFs},'' \emph{TPAMI}, 2017.

\bibitem{Chen17_3}
Q.~Chen and V.~Koltun, ``Photographic image synthesis with cascaded refinement
  networks,'' \emph{{IEEE} International Conference on Computer Vision}, 2017.

\bibitem{Yang18}
B.~Yang, W.~Luo, and R.~Urtasun, ``{PIXOR}: Real-time {3D} object detection
  from point clouds,'' \emph{IEEE Conference on Computer Vision and Pattern
  Recognition}, 2018.

\bibitem{Parmar18}
N.~Parmar, A.~Vaswani, J.~Uszkoreit, L.~Kaiser, N.~Shazeer, A.~Ku, and D.~Tran,
  ``Image transformer,'' \emph{ICML}, 2018.

\bibitem{Brock19}
A.~Brock, J.~Donahue, and K.~Simonyan, ``Large scale gan training for high
  fidelity natural image synthesis,'' \emph{ICLR}, 2019.

\bibitem{Mildenhall2020}
B.~Mildenhall, P.~P. Srinivasan, M.~Tancik, J.~T. Barron, R.~Ramamoorthi, and
  R.~Ng, ``{NeRF}: Representing scenes as neural radiance fields for view
  synthesis,'' \emph{ECCV}, 2020.

\bibitem{Park20}
T.~Park, A.~A. Efros, R.~Zhang, and J.~Zhu, ``Contrastive learning for unpaired
  image-to-image translation,'' \emph{EUROPEAN Conference on Computer Vision},
  2020.

\bibitem{Odena17}
A.~Odena, C.~Olah, and J.~Shlens, ``Conditional image synthesis with auxiliary
  classifier gans,'' \emph{ICML}, 2017.

\bibitem{Antoniou18}
A.~Antoniou, A.~Storkey, and H.~Edwards, ``Data augmentation generative
  adversarial networks,'' \emph{ICLR}, 2018.

\bibitem{Salimans16}
T.~Salimans, I.~Goodfellow, W.~Zaremba, V.~Cheung, A.~Radford, and X.~Chen,
  ``Improved techniques for training {GANs},'' \emph{NIPS}, 2016.

\bibitem{Mescheder17}
L.~Mescheder, S.~Nowozin, and A.~Geiger, ``The numerics of {GANs},''
  \emph{NIPS}, 2017.

\bibitem{Mescheder18}
------, ``Which training methods for {GANs} do actually converge?''
  \emph{ICML}, 2018.

\bibitem{Jolicoeur2019}
A.~{Jolicoeur-Martineau}, ``The relativistic discriminator: a key element
  missing from standard {GAN},'' \emph{ICLR}, 2019.

\bibitem{Karras2019}
T.~Karras, S.~Laine, and T.~Aila, ``A style-based generator architecture for
  generative adversarial networks,'' \emph{CVPR}, 2019.

\bibitem{Karras2021}
T.~Karras, M.~Aittala, S.~Laine, E.~H\"ark\"onen, J.~Hellsten, J.~Lehtinen, and
  T.~Aila, ``Alias-free generative adversarial networks,'' \emph{NIPS}, 2021.

\bibitem{Kingma14}
D.~P. Kingma and M.~Welling, ``Auto-encoding variational bayes,'' \emph{ICLR},
  2014.

\bibitem{Dosovitskiy16}
A.~Dosovitskiy and T.~Brox, ``Generating images with perceptual similarity
  metrics based on deep networks,'' \emph{NIPS}, 2016.

\bibitem{Zhao17}
J.~Zhao, M.~Mathieu, and Y.~LeCun, ``Energy-based generative adversarial
  networks,'' \emph{ICLR}, 2017.

\bibitem{Wei18}
X.~Wei, B.~Gong, Z.~Liu, W.~Lu, and L.~Wang, ``Improving the improved training
  of wasserstein gans: a consistency term and its dual effect,'' \emph{ICLR},
  2018.

\bibitem{Song2019}
Y.~Song and S.~Ermon, ``Generative modeling by estimating gradients of the data
  distribution,'' \emph{NIPS}, 2019.

\bibitem{Roberts1996}
G.~O. Roberts and R.~L. Tweedie, ``{Exponential convergence of Langevin
  distributions and their discrete approximations},'' \emph{Bernoulli}, vol.~2,
  no.~4, pp. 341 -- 363, 1996.

\bibitem{Welling2011}
M.~Welling and Y.~W. Teh, ``Bayesian learning via stochastic gradient langevin
  dynamics,'' \emph{ICML}, 2011.

\bibitem{Song2020}
Y.~Song and S.~Ermon, ``Improved techniques for training score-based generative
  models,'' \emph{NIPS}, 2020.

\bibitem{Ho2020}
J.~Ho, A.~Jain, and P.~Abbeel, ``Denoising diffusion probabilistic models,''
  \emph{NIPS}, 2020.

\bibitem{Goyal2017}
A.~Goyal, N.~R. Ke, S.~Ganguli, and Y.~Bengio, ``Variational walkback: Learning
  a transition operator as a stochastic recurrent net,'' \emph{NIPS}, 2017.

\bibitem{Jolicoeur2021}
A.~Jolicoeur{-}Martineau, R.~Pich{\'{e}}{-}Taillefer, R.~T. {des Combes}, and
  I.~Mitliagkas, ``Adversarial score matching and improved sampling for image
  generation,'' \emph{ICLR}, 2021.

\bibitem{Zhao2021}
L.~Zhao, Z.~Zhang, T.~Chen, D.~N. Metaxas, and H.~Zhang, ``Improved transformer
  for high-resolution gans,'' \emph{NIPS}, 2021.

\bibitem{Dosovitskiy2020}
A.~Dosovitskiy, L.~Beyer, A.~Kolesnikov, D.~Weissenborn, X.~Zhai,
  T.~Unterthiner, M.~Dehghani, M.~Minderer, G.~Heigold, S.~Gelly, J.~Uszkoreit,
  and N.~Houlsby, ``An image is worth 16x16 words: Transformers for image
  recognition at scale,'' \emph{ICLR}, 2020.

\bibitem{vandermaaten08a}
L.~{van der Maaten} and G.~Hinton, ``Visualizing data using {t-SNE},''
  \emph{JMLR}, vol.~9, no.~86, pp. 2579--2605, 2008.

\bibitem{Karras2020}
T.~Karras, M.~Aittala, J.~Hellsten, S.~Laine, J.~Lehtinen, and T.~Aila,
  ``Training generative adversarial networks with limited data,'' \emph{NIPS},
  2020.

\bibitem{KruskalandWallis52}
W.~H. Kruskal and W.~A. Wallis, ``Use of ranks in one-criterion variance
  analysis,'' \emph{Journal of the American Statistical Association}, vol.~47,
  no. 260, pp. 583--621, 1952.

\bibitem{Choi2020}
Y.~Choi, Y.~Uh, J.~Yoo, and J.~Ha, ``Stargan v2: Diverse image synthesis for
  multiple domains,'' \emph{CVPR}, 2020.

\bibitem{Bonnier2019}
P.~Bonnier, P.~Kidger, I.~P. Arribas, C.~Salvi, and T.~Lyons, ``Deep signature
  transforms,'' \emph{NIPS}, 2019.

\bibitem{Kidger2021}
P.~Kidger and T.~Lyons, ``{S}ignatory: differentiable computations of the
  signature and logsignature transforms, on both {CPU} and {GPU},''
  \emph{ICLR}, 2021.

\bibitem{Chevyrev2016}
I.~Chevyrev and A.~Kormilitzin, ``A primer on the signature method in machine
  learning,'' 2016.

\bibitem{Liao2019}
S.~Liao, T.~J. Lyons, W.~Yang, and H.~Ni, ``Learning stochastic differential
  equations using {RNN} with log signature features,'' \emph{arXiv:1908.08286},
  2019.

\bibitem{Morrill2021}
J.~Morrill, P.~Kidger, C.~Salvi, J.~Foster, and T.~J. Lyons, ``Neural {CDEs}
  for long time series via the log-ode method,'' \emph{ICML}, 2021.

\bibitem{Kiraly2019}
F.~J. Kiraly and H.~Oberhauser, ``Kernels for sequentially ordered data,''
  \emph{JMLR}, vol.~20, no.~31, pp. 1--45, 2019.

\bibitem{Graham2013}
B.~Graham, ``Sparse arrays of signatures for online character recognition,''
  2013.

\bibitem{Chang2019}
J.~Chang and T.~Lyons, ``Insertion algorithm for inverting the signature of a
  path,'' \emph{arXiv:1907.08423}, 2019.

\bibitem{Fermanian21}
\BIBentryALTinterwordspacing
A.~Fermanian, ``{Learning time-dependent data with the signature transform},''
  Theses, {Sorbonne Universit{\'e}}, 2021. [Online]. Available:
  \url{https://tel.archives-ouvertes.fr/tel-03507274}
\BIBentrySTDinterwordspacing

\bibitem{Lyons1998}
T.~Lyons, ``Differential equations driven by rough signals,'' \emph{Revista
  Matem{\'a}tica Iberoamericana}, vol.~14, no.~2, pp. 215--310, 1998.

\bibitem{Lyons2007}
T.~Lyons, M.~Caruana, and T.~L{\'e}vy, ``Differential equations driven by rough
  paths,'' \emph{Springer}, 2007.

\bibitem{DeCurto22_3}
J.~{de Curtò}, I.~{de Zarzà}, H.~Yan, and C.~T. Calafate, ``On the
  applicability of the hadamard as an input modulator for problems of
  classification,'' \emph{Software Impacts}, vol.~13, p. 100325, 2022.

\end{thebibliography}

\end{document}